\documentclass[a4paper]{llncs}
\usepackage{graphicx}
\usepackage{makeidx}  
\usepackage{url}
\usepackage{footnote}
\usepackage{etoolbox} 
\usepackage{enumitem}
\makeatletter
\patchcmd{\ps@headings}{\rlap{\thepage}}{}{}{}
\patchcmd{\ps@headings}{\llap{\thepage}}{}{}{}
\makeatother
\usepackage{hyperref}

\makeatletter
\newcommand{\printfnsymbol}[1]{%
  \textsuperscript{\@fnsymbol{#1}}%
}
\makeatother
\usepackage{mathtools}
\usepackage{amsmath}
\usepackage{lipsum}
\makeatletter
\def\blfootnote{\gdef\@thefnmark{}\@footnotetext}
\makeatother
\begin{document}
%
%

%
\mainmatter              
\title{Deep Q-Network for AI Soccer}
\blfootnote{This work was supported by the Institute for Information \& Communications Technology Planning and Evaluation (IITP) grant funded by the Korea government (MSIT) (No.2020-0-00440, Development of Artificial Intelligence Technology that Continuously Improves Itself as the Situation Changes in the Real World.}
\titlerunning{Hamiltonian Mechanics}  
%
\author{Curie Kim\inst{1}\thanks{These authors equally contributed to this work} \and Yewon Hwang\inst{2}\printfnsymbol{1} \and Jong-Hwan Kim\inst{2}}

%
%
%
\institute{Samsung Electronics\\
\and KAIST 
} 

\maketitle              

\begin{abstract}

Reinforcement learning has shown an outstanding performance in the applications of games, particularly in Atari games as well as Go. Based on these successful examples, we attempt to apply one of the well-known reinforcement learning algorithms, Deep Q-Network, to the AI Soccer game. AI Soccer is a 5:5 robot soccer game where each participant develops an algorithm that controls five robots in a team to defeat the opponent participant. Deep Q-Network is designed to implement our original rewards, the state space, and the action space to train each agent so that it can take proper actions in different situations during the game. Our algorithm was able to successfully train the agents, and its performance was preliminarily proven through the mini-competition against 10 teams wishing to take part in the AI Soccer international competition. The competition was organized by the AI World Cup committee, in conjunction with the WCG 2019 Xi'an AI Masters. With our algorithm, we got the achievement of advancing to the round of 16 in this international competition with 130 teams from 39 countries.
\end{abstract}
\section{Introduction}
Reinforcement learning is one of the most studied domains in machine learning thanks to its drastic improvement over the past few years. Reinforcement learning has led to recent breakthroughs of machine learning applications in games, prominently in Atari games \cite{atari}, and more recently, the Chinese game of Go \cite{Silver_2016}. What differs reinforcement learning from traditional machine learning is that there is no training dataset or labels, and agents are to learn how to behave in an environment by performing actions and observing rewards that it receives, where the goal of the agent is to find a set of actions that will maximize the total reward. Essentially, the idea of reinforcement learning is that agents are to learn based on their experience just like how humans learn. Since reinforcement learning enables an agent to learn autonomously from its own experience, it is a great framework for learning behaviors of agents in applications such as soccer games, in particular, where a vast amount of possibilities can occur.

Efforts to use reinforcement learning in various games have been made
\cite{StarCraft}\cite{Dota}\cite{ViZDoom}. The AI World Cup, a set of AI competitions based on the game of soccer, was even established in 2017 and the official international AI World Cup has been held in 2018 and 2019 and the AI Masters competition was held in 2019 as a part of the World Cyber Games \cite{aiwc}. Some organizations also have held soccer simulation tournaments \cite{fira}\cite{robocup}. In addition, various strategies and methods for obtaining good results in the tournaments were presented \cite{inbook}\cite{Fukushima}. Among them, we used the AI Soccer environment proposed by \cite{aiwc} (available at \cite{git}) for the experiment. AI Soccer is a 5:5 robot soccer game where each team is composed of one goalkeeper, two defenders, and two forwards. However, the role does not affect the robots’ capabilities: the only difference between each player is its specifications and initial position. The game is comprised of 5 minutes first half and 5 minutes second half. At the beginning of each half and after a team makes a goal, kick-off happens where only the second forward player of the ball owner’s team can move. 

\begin{figure}
    \centerline{\includegraphics[width=10cm]{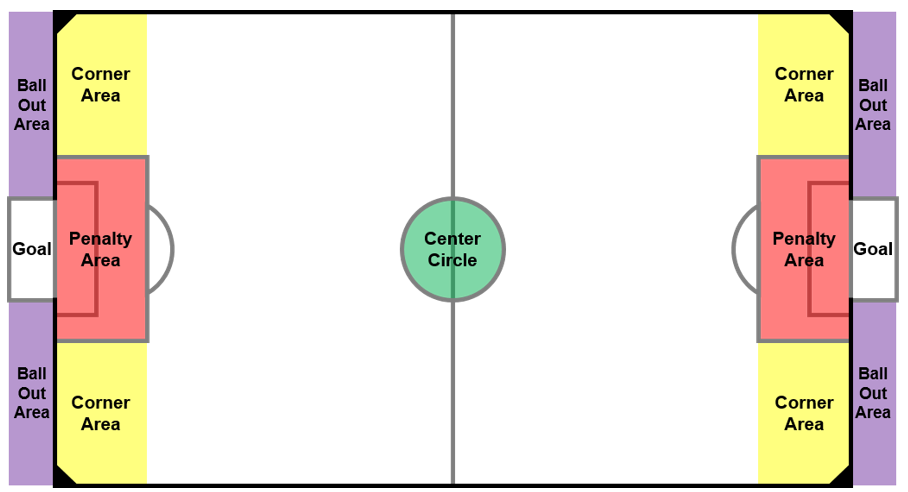}}
    \caption[Different areas on the soccer field]{Different areas in the soccer field. The figure is taken from \cite{aiworldcup}.
    } \label{areas}
\end{figure}

In AI Soccer, there is a situation where the robots fail to track the soccer ball correctly, preventing the game to proceed further. In order to avoid and handle such situations effectively, three different deadlock rules are implemented. Deadlock here is referred to as a situation where the soccer ball moves at a speed that is slower than 0.4 m/s for 4 seconds. First, deadlock can occur in one of the four corner areas. In such a case, the game proceeds to a corner kick. A corner kick can also be initiated when the ball leaves the soccer field, and no robots except the second forward player in the ball owner’s team can move. Secondly, deadlock can happen in one of the two penalty areas, then the game proceeds either to a penalty kick if the ball owner is on the opposite side of their goalpost or a goal kick, otherwise. In such a situation, if the foul is made by the defense team, the defending team can only have three robots inside the penalty area at the same time, while if the foul is made by the offense team, the attacking team can only have two robots inside the penalty area at the same time. Lastly, deadlock can occur in other regions which will lead to a ball relocation. There are four ball relocation positions in total, and during ball relocation, the ball will be relocated to a position closest to the current ball. Each region of the soccer field is shown in Fig. \ref{areas}.

There is a situation where a robot falls and fails to recover for more than 3 seconds. If such a case happens, the robot is moved out of the soccer field and stays inactive for 5 seconds. After 5 seconds, the robot is moved back to the designated position and orientation on the soccer field. 

Reinforcement learning can be explained using numerous factors such as an agent, actions, environment, states, and rewards. An agent is the one that takes an action, where the action is the set of all possible behavior that the agent can take. The environment is the world that the agent is navigating through, and the state is a situation in which the agent finds itself. The Reward is the feedback that the agent receives indicating how good or bad the action was. The strategy that the agent utilizes to determine the next action is called policy, $\pi$. A value function, \textit{$V_{\pi}(s)$}, estimates how good it is for the agent to be in a given state, and it is expressed as follows:
\begin{eqnarray}
V_{\pi}(s)=E_{\pi}\left[\sum_{k=0}^{\infty} \gamma^{k} R_{t+k+1} \mid S_{t}=s\right]
\label{eqn1}
\end{eqnarray}

Similarly, the long-term return of the current state, $s$, taking action, $a$, under the policy, $\pi$, is referred to as the action-value function, $Q_\pi(s, a)$, expressed as follows:
\begin{eqnarray}
Q_{\pi}(s, a)=E_{\pi}\left[\sum_{k=0}^{\infty} \gamma^{k} R_{t+k+1} \mid S_{t}=s, A_{t}=a\right].
\label{eqn2}
\end{eqnarray}

Many reinforcement learning algorithms follow Markov Decision Process (MDP) to approximate the probability distribution of reward over state-action pairs and the idea of estimating the action-value function by using the Bellman equation. MDP is a mathematical framework to describe an environment in reinforcement learning and almost all RL problems can be formalized using MDPs. An MDP consists of a set of states $S$, a set of possible actions $A(s)$ in each state, a reward function $R(s)$, and a transition model $P(s\prime, s | a)$. The above two value equations obey the Bellman equation, described in equation \ref{eqn3}, of which the fundamental idea is that if the optimal value of the sequence at the next time step is known for all possible actions, then the optimal strategy is to select the action which maximizes the expected value of return \cite{atari} as follows:
\begin{eqnarray}
Q(s, a)=r(s, a)+\gamma \max _{a} Q\left(s^{\prime}, a\right).
\label{eqn3}
\end{eqnarray}

Reinforcement learning algorithms repeat actions that lead to reward, which is called exploitation. However, this will stop agents to take exploration, which could eventually lead to a better reward, and lead to a narrow number of paths agents can take. In order to prevent agents from only exploiting, Epsilon Greedy Policy is employed, where agents are to take a known path with epsilon probability and explore the rest.
Many researchers have proposed different algorithms for reinforcement learning such as Deep Q-Network (DQN), Proximal Policy Optimization (PPO) \cite{ppo}, Deep Deterministic Policy Gradient (DDPG), and Soft Actor-Critic (SAC) \cite{sac}, etc. In order to train our agents for the AI Soccer game, we employed the DQN algorithm.

\section{Related Work}

In this section, we review some of the most well-known reinforcement learning algorithms that we attempted to use for AI Soccer to train our agents.

\subsection{Deep Q-Network (DQN)}
Deep Q-Network \cite{atari} was first proposed in 2013 demonstrating its superior performance on seven Atari 2600 games. DQN is the first deep learning model to successfully learn control policies directly from high-dimensional input using reinforcement learning. The model is a convolutional neural network whose input is raw pixels and output is a value function estimating future rewards. The network is trained with a variant of the Q-learning algorithm with stochastic gradient descent to update the weights. 

In reinforcement learning, both the input and the target change constantly during the training process and make training unstable. To resolve instability, DQN employs two networks, one for retrieving Q-values, while the other one includes all updates in the training. After a set number of updates, the two networks are synchronized to fix the parameters of the target function and replace them with the latest network. In addition, DQN adapts the experience replay mechanism which randomly samples previous transitions and forms an input dataset with enough stability for training. To alleviate the high correlation between data and to make data independent of each other, data are then randomly sampled from the replay buffer. Lastly, to obtain the optimal action-value function, the following equation is used:
\begin{eqnarray}
Q^{*}(s, a)=E_{s^{\prime} \sim \varepsilon}\left[r+\gamma \max _{a} Q^{*}\left(s^{\prime}, a^{\prime}\right) \mid s, a\right].
\label{eqn4}
\end{eqnarray}

While DQN solves problems with high-dimensional observation spaces, it can only handle discrete and low-dimensional action spaces. Hence, to address the limitations of DQN, Deep Deterministic Policy Gradient (DDPG) was introduced.

\subsection{Deep Deterministic Policy Gradient (DDPG)}
Deep Deterministic Policy Gradient (DDPG) \cite{ddpg} was first introduced in 2015 to address the limitations of DQN. DDPG is an actor-critic, model-free algorithm that can operate over continuous action spaces. The algorithm can solve more than 20 simulated tasks such as cart-pole, dexterous manipulation, legged locomotion, and car driving.
DDPG is an algorithm that concurrently learns a Q-function and a policy. It uses off-policy data and the Bellman equation to learn the Q-function and uses the Q-function to learn the policy. DDPG utilizes Q-learning, of which fundamental idea is that if you know the optimal action-value function, then in any given state, the optimal action can be found by solving the following equation:
\begin{eqnarray}
a^{*}(s)=\arg \max _{a} Q^{*}(s, a).
\label{eqn5}
\end{eqnarray}

The optimal action-value function can be found using the same equation \ref{eqn4} used in DQN. 

DDPG is comprised of two networks: the actor and the critic network. The actor function, $\mu(s | \theta_{\mu})$, in the actor network specifies action given the current state of the environment. The critic value function, $Q(s, a | \theta_{Q})$, in the critic network then calculates the TD error to criticize the actions made by the actor. Similar to DQN, DDPG ensures exploration by introducing noise to the action, and it also uses the same experience replay mechanism adopted in DQN to have a stable behavior. 
\section{Proposed Method and Strategy}
For training the agents, we initially attempted to adopt the Deep Deterministic Policy Gradient algorithm. However, one critical issue that we encountered during the training process using DDPG was the position values of each agent were diverging to either -1 or 1 because the computation time for calculating the action and reward was taking longer than the time it takes to receive the next frame from the Webots platform. Since the two times were not in sync, neither the calculated action nor the reward from the network was reflected during the training, preventing a proper learning process. Hence, consequently, we used Deep Q-Network to train defense and forward players and used a rule-based scheme for the goalkeeper.

\subsection{State Space}
For state space, we define 22 states to train the agents. The robots’ coordinate and orientation values as well as the ball’s positions and orientations were provided through Webots throughout the game. All the positions were provided in meters in a Cartesian coordinate system, and the orientations were provided in radian. We first defined the x, y, and $\theta$ values of each player excluding the goalkeeper, and applied a regularization process to each value to avoid the risk of overfitting. Since there are two defense players and two forward players, 12 states are defined thus far. We also provided boolean information that indicates whether a player is active or inactive, in case a player is dismissed due to varying reasons for both forward and defense players as a state, which makes 16 states in total thus far. In addition, we provided 4 more states which are two x and y values of the ball that have undergone the regularization process. The reason we gave each x and y value twice is to give more weight to the values because we consider that information about the ball was the most critical information when training the agents. Lastly, we also provided x and y values of the ball after two frames from the current frame that have undergone the regularization process. The motivation behind the last two states is to train the agents so that they can predict the position of the ball in the next two frames and act accordingly in favor of our team. Therefore, there are 22 states in total as follows:

\begin{equation}
\label{eq:1}
\begin{aligned}
s = &\{ 4_{(F1,F2, D1, D2)} \times (player\_x, player\_y, player\_\theta, is\_active), \\ &2 \times (ball\_x, ball\_y), (predicted\_ball\_x, predicted\_ball\_y) \}.
\end{aligned}
\end{equation}

\subsection{Action Space}
For action space, we provided 256 actions in total. First, based on the current speed of the ball, we predicted the position of the ball in the next two frames using \texttt{predict\_ball\_location} function provided in the rule-based example code by \cite{aiworldcup}. Based on the ball’s location, we set each agent’s action except the goalkeeper so that each of them moves to the \textit{above}, \textit{below}, \textit{left}, and \textit{right} of the ball. Therefore, each agent except the goalkeeper can have 4 target positions. The motivation behind setting the actions this way, in particular, is to train the agents so that they can successfully follow the ball. By following the ball properly, we believe that there is a greater chance that one of the agents will score a goal. Since one-hot encoding must be used to train Deep Q-Network, four agents have values for all four target positions. For example, if the target position is encoded in the order of [\textit{above}, \textit{below}, \textit{left}, \textit{right}] and the target position of \textit{forward 1} is [0, 1, 0, 0], this indicates that the action of \textit{forward 1} is \textit{below}.
Since each of the four agents can choose one of four actions independently of each other, the team as a whole can have 256 different action combinations. Our network chooses the best one out of 256 actions, and the positions of the four agents are determined automatically according to the order of the encoded target positions.

\subsection{Reward Signal}
\begin{figure}[hb!]    \centerline{\includegraphics[width=10cm]{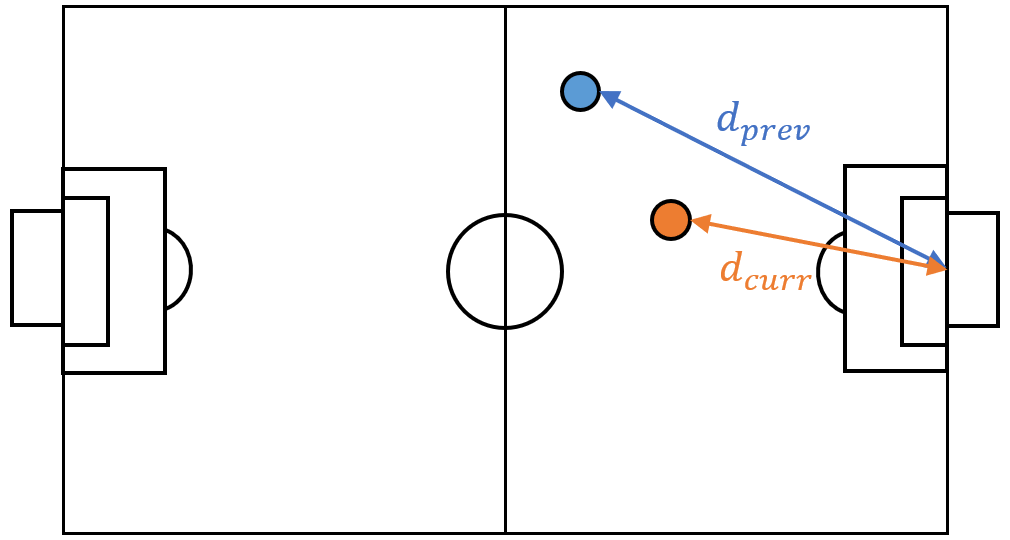}}
    \caption[dcurr]{Distances between the opponent's goalpost and the ball of the current and the previous frame.
    } \label{dcurr}
\end{figure}

We defined six reward signals in total based on the region the robots are located. We used the same reward function for all six instances but used different parameters for each instance. We describe the reward function and each of the six rewards as follows:
\begin{eqnarray}
{Reward}=\mathrm{C}_{1}+\mathrm{C}_{2} \times\left(\mathrm{d}_{prev}-\mathrm{d}_{curr}\right),
\label{eqn6}
\end{eqnarray}
 where $d_{prev}$ indicates the distance between the opposing team’s goalpost and the ball of the second previous frame and $d_{curr}$ indicates a current distance between the opponent's goalpost and the ball (see Fig. \ref{dcurr}). 

Below are our rules that determine rewards according to separated regions (see Table \ref{params}).

\begin{figure}[]
    \centerline{\includegraphics[width=10cm]{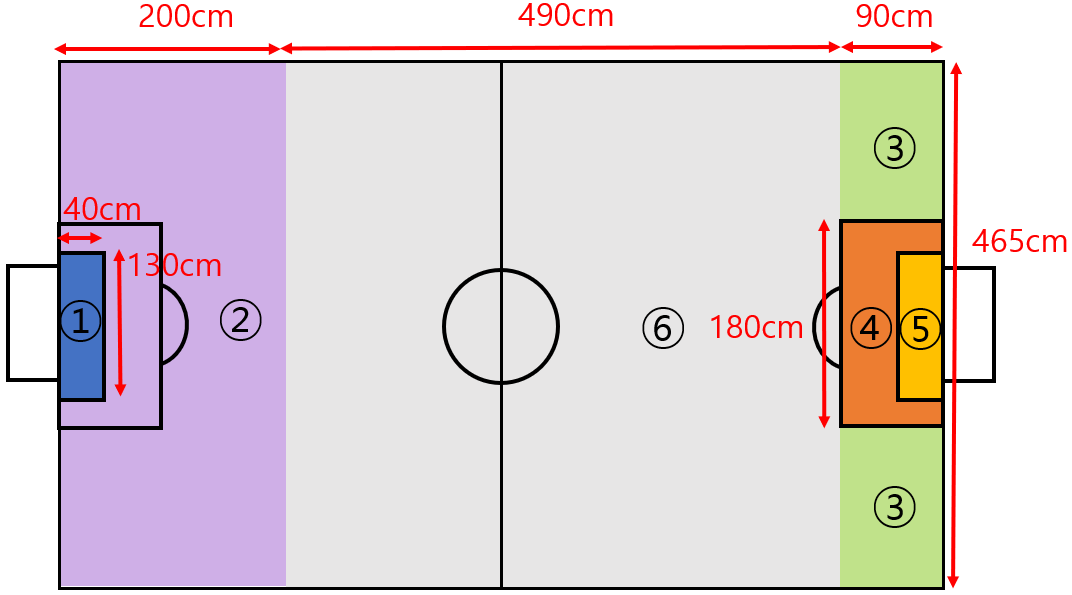}}
    \caption[region]{Different regions in the soccer field for granting reward values.
    } \label{region}
\end{figure}

 \begin{itemize}
  \item If the agents are near the opponent’s goalpost, where it is shaded in yellow and indicated as \textcircled{\small 5} in Figure \ref{region}, we set $C_1$ to 10 and $C_2$ to 0, granting each agent positive 10 rewards.
  \item If the agents are near our goal post, where it is shaded in blue and indicated as \textcircled{\small 1} in Figure \ref{region}, we set $C_1$ to -10 and $C_2$ to 0, granting each agent negative 10 rewards.
  \item In the penalty area shaded in orange and indicated as \textcircled{\small 4}, we set $C_1$ to 1 and $C_2$ to 10.
  \item In the corner areas in the opponent’s region which are shaded in green and indicated as \textcircled{\small 3}, we set $C_1$ to 0.5 and $C_2$ to 10.
  \item If the agents are in our region, where it is shaded in purple and indicated as \textcircled{\small 2} in Figure \ref{region}, we set $C_1$ to -1 and $C_2$ to 10.
  \item In the rest of the region shaded in gray and indicated as \textcircled{\small 6}, we set $C_1$ to 0 and $C_2$ to 10.
\end{itemize}

The motivation behind the first reward signal is that since we train our agents to follow the ball if the agents, as well as the ball, are near the opponent’s goalpost, it is likely that our agent will score a goal. Therefore, in order to encourage such an instance to happen, we give the agent a high positive reward. Similarly, the incentive behind the second reward signal is that if our agents, as well as the ball, are near our team’s goalpost, it is more likely that our agent will score its own goal. Hence, in order to prevent such occasions, we give the agent a high negative reward value. The motivation behind the rest of the rewards is that since the ball moving closer to the opponent’s goalpost area is advantageous for us, if the distance between the ball and the opponent’s goal post has reduced from the previous frame to the current frame, we reward each agent by multiplying 10 to the difference. Moreover, based on the region, we add different values. Since it is more likely to score a goal if the ball, as well as the players, are near the opponent’s goalpost, in regions \textcircled{\small 3} and \textcircled{\small 4} which are closer to the opponent’s goalpost, we add 1 and 0.5, respectively. On the other hand, in region \textcircled{\small 2}, we add -1 since we want to penalize a situation where the ball and the players are near our goalpost. Finally, since region \textcircled{\small 6} is a neutral zone, we do not add any value as a reward. 

\begin{table}
\caption{$C_1$ and $C_2$ parameters for different regions.}
\begin{center}
\begin{tabular}{ccc}
\hline\rule{0pt}{12pt}
\textbf{Region  }\hspace{5mm}  & \textbf{$C_1$ parameter}\hspace{5mm}   & \textbf{$C_2$ parameter} \\ [2pt]
\hline
1               & -10                      & 0                       \\ 
2               & -1                       & 10                      \\ 
3               & 0.5                      & 10                      \\ 
4               & 1                        & 10                      \\ 
5               & 10                       & 0                       \\ 
6               & 0                        & 10                      \\ \hline
\end{tabular}
\end{center}
\label{params}
\end{table}

\subsection{Deep Q-Network Architecture}
Deep Q-Network is composed of two networks: a behavior network and a target network. A behavior network determines the action based on Q-value that has been trained. During a training process, 22 states, one action, and one reward are stored in a replay memory every frame. Once the total number of values that are stored in the replay memory becomes greater or equal to 5,000, a minibatch with a size of 64 is randomly sampled for training. A target network periodically copies the behavior network and conducts a learning process. The behavior network is trained by reducing the difference between the Q-value calculated by the behavior network using the minibatch of a frame and the Q-value calculated by the target network using the minibatch of the next frame by using gradient descent. At the beginning of the training, we allowed the agents to take exploration using Epsilon greedy policy. This way, agents are to take a random action with a probability of epsilon, rather than taking an action determined by the Deep Q-Network. We set the value of epsilon as 1 at the beginning of the training and reduced the value by 0.05 every 20,000 times of training. 

Our Deep Q-Network consists of an input layer, two hidden layers, and an output layer. The input layer and output layer have 22 and 256 nodes, respectively, since there are 22 state spaces and 256 action spaces. Both hidden layers have 256 nodes. We adopted Rectified Linear Unit as our activation function and optimized our network using the Adam optimizer. In total, 15 hours were spent on training, and we stopped the training when the Q-value loss reached the minimum.

\subsection{Rule-based Scheme for Goalkeeper}
We used a rule-based scheme for the goalkeeper, rather than using the learning-based one because we thought that the rule-based would be more effective for goalkeepers since there is less number of situations that goalkeepers have to handle. We employed numerous rules according to different conditions, and we describe each rule-based strategy we used for the goalkeeper below:
 \begin{enumerate}
  \item During the \texttt{default state}:
    \begin{itemize}
        \item y-position of the goalkeeper is set in line with the y-position of the ball
        \item If the goalkeeper is inside the goal, it tries to get out.
        \item If the goalkeeper is outside the penalty area, it returns to the desired position.
    \end{itemize}
  \item When the goalkeeper, as well as the ball, are inside the penalty area:
    \begin{itemize}
        \item If the ball is behind the goalkeeper and is not blocking the goalkeeper's path, it tries to get ahead of the ball.
        \item If not, it gives up and tries to avoid making its own goal. 
    \end{itemize}
  \item When the goalkeeper, as well as the ball, are inside the penalty area:
    \begin{itemize}
        \item If the direction of the robot is far away from the ball direction, it gives up kicking the ball and blocks the goalpost.
        \item If not, it tries to kick the ball away from the goalpost.If the ball is within an alert range, and there is not much difference in the y-position of the ball and the goalkeeper, the goalkeeper gazes at the ball.
        \item If not, it goes to the desired position.
    \end{itemize}
  \item When the goalkeeper is inside the penalty area, but the ball is not in the penalty area:
    \begin{itemize}
        \item If the ball is within an alert range, and there is not much difference in the y-position of the ball and the goalkeeper, the goalkeeper gazes at the ball.
        \item If not, it goes to the desired position.
    \end{itemize}

\end{enumerate}

\subsection{Different strategies for different states}

\begin{figure}
    \centerline{\includegraphics[width=10cm]{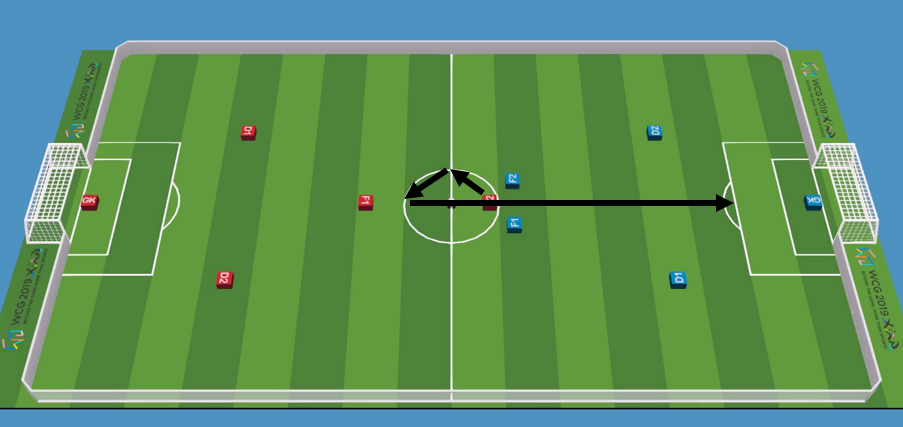}}
    \caption[kickoff]{Position of \textit{forward 2} each frame during a \texttt{kick-off}.
    } \label{kickoff}
\end{figure}

We adopted different strategies for different states such as \texttt{default state}, \texttt{kick-off}, \texttt{goal kick}, \texttt{corner kick}, and \texttt{penalty kick}. 
\begin{figure}[t]
    \centerline{\includegraphics[width=10cm]{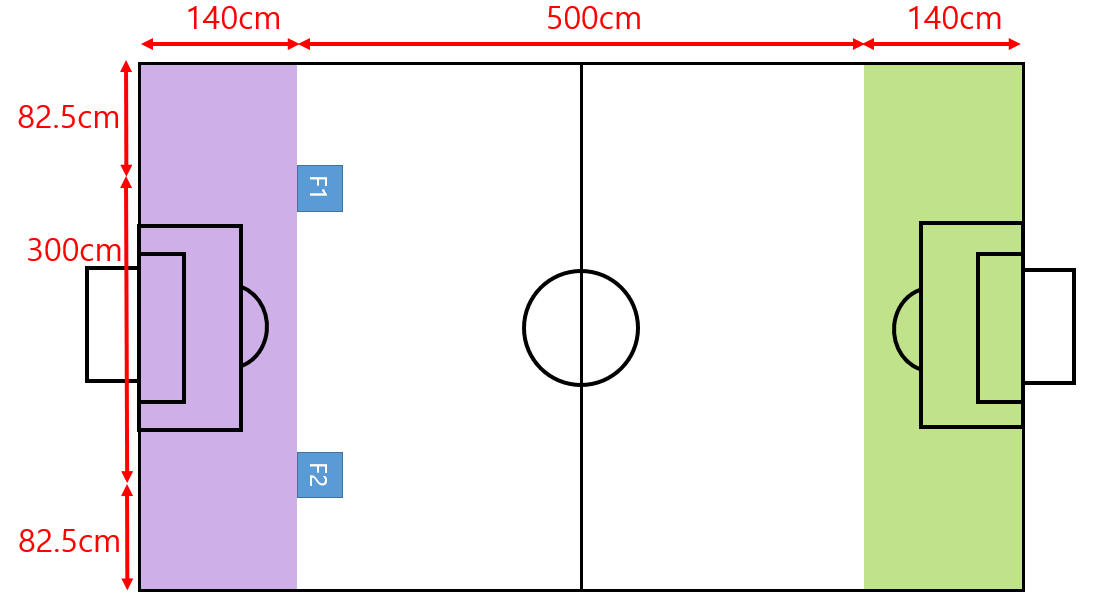}}
    \caption[teamregion]{Two different regions where the purple area indicates our team’s region, while the green area indicates the opponent’s region.
    } \label{teamregion}
\end{figure}First, during the \texttt{default state}, to prevent a penalty kick, two defense players are not allowed inside the green area in Figure \ref{teamregion}, but they are to only follow the y-position of their action. Similarly, two forward players are not allowed inside the purple area in the figure, but they are to only follow the y-position of their action. Also, to prevent an own goal, if the ball is inside the purple area, \textit{defender 1} tries to match the y-position of the goalkeeper to help defend. \textit{Forward 1} and \textit{forward 2} tend to cause an own goal; thus, to prevent an own goal, they are to wait in a position indicated in the figure until the ball gets out of the purple region. Secondly, we set a strategy for scoring a goal by taking advantage of the fact that only \textit{forward 2} is able to move during the kick-off. We made \textit{forward 2} go around the ball and shoot as shown in Figure \ref{kickoff} by setting the position of the robot in each frame using the \texttt{set\_target\_position} function provided in the example code. Thirdly, during the \texttt{goal kick}, we set both wheels of the goalkeeper at a maximum velocity so that the ball moves toward the opponent’s region as far as possible. Fourthly, during the \texttt{corner kick}, \textit{forward 2} is to move to the left of the ball and shoot so that the ball moves to the opponent’s goalpost as close as possible. Lastly, during the \texttt{penalty kick}, if our team has ball ownership, the kicker shoots the ball aiming slightly above the center line, rather than the center to make it harder for the opponent’s goalkeeper to block the ball.

\section{Results and Conclusion}

Our algorithm was preliminarily tested through the mini-competition against 10 teams of participants wishing to take part in the AI Soccer international competition, organized by the AI World Cup committee, in conjunction with the WCG 2019 Xi'an AI Masters \cite{wcg}. 9 out of 10 algorithms were learning-based and only one algorithm was rule-based. However, the details of others' algorithms have not been made public. Table \ref{tb:result} summarizes the result of the preliminary mini-competition. Our team won against 7 teams, lost against 2 teams, and tied with one team, resulting in our team being placed second among the entire teams. After preliminary testing against 10 teams, at last, we participated in the AI Soccer international competition with 130 teams from 39 countries. We advanced to the round of 16. Our team name is CY95, and the recorded video of the round of 16 is available at \cite{Youtube}.

\begin{table}
\caption{Results of the preliminary mini-competition.}
\begin{center}
\begin{tabular}{ccc}
\hline\rule{0pt}{12pt}
\textbf{Opponent  }\hspace{5mm}  & \textbf{Score}\hspace{5mm}   & \textbf{Victory} \\ [2pt]
\hline
Ours vs Team 1               & 13:8                 & Win                  \\ 
Ours vs Team 2               & 10:8                 & Win                  \\ 
Ours vs Team 3               & 7:13                 & Lose                  \\ 
Ours vs Team 4               & 6:6                 & Tie                  \\
Ours vs Team 5               & 8:7                 & Win                  \\
Ours vs Team 6               & 5:1                 & Win                  \\ 
Ours vs Team 7               & 20:3                 & Win                  \\ 
Ours vs Team 8               & 23:6                 & Win                  \\ 
Ours vs Team 9               & 8:10                 & Lose                  \\ 
Ours vs Team 10               & 8:4                 & Win                  \\ \hline

\end{tabular}
\end{center}
\label{tb:result}
\end{table}
Deep Q-Network was applied for training our agents in AI Soccer. We implemented DQN to train each agent so that it can take proper actions in different situations. We set the state space and the action space and rewards optimized for the AI Soccer task in a creative own way, and our algorithm was able to successfully train the agents. The performance of our algorithm was proven by 10 matches against 10 teams. Our algorithm proved its performance by advancing to the round of 16 in the AI Soccer international competition among 130 teams from 39 countries, organized by the AI World Cup committee, in conjunction with the WCG 2019 Xi'an AI Masters by successfully applying DQN to AI Soccer.

%
%
\bibliographystyle{splncs}
\bibliography{reference} 

\begin{thebibliography}{10}

\bibitem{atari}
Mnih, V., Kavukcuoglu, K., Silver, D., Graves, A., Antonoglou, I., Wierstra,
  D., Riedmiller, M.:
\newblock Playing atari with deep reinforcement learning (2013)

\bibitem{Silver_2016}
Silver, D., Huang, A., Maddison, C.J., Guez, A., Sifre, L., van~den Driessche,
  G., Schrittwieser, J., Antonoglou, I., Panneershelvam, V., Lanctot, M.,
  Dieleman, S., Grewe, D., Nham, J., Kalchbrenner, N., Sutskever, I.,
  Lillicrap, T., Leach, M., Kavukcuoglu, K., Graepel, T., Hassabis, D.:
\newblock Mastering the game of {Go} with deep neural networks and tree search.
\newblock Nature \textbf{529}(7587) (January 2016)  484--489

\bibitem{StarCraft}
Čertický, M., Churchill, D., Kim, K.J., Čertický, M., Kelly, R.:
\newblock Starcraft ai competitions, bots, and tournament manager software.
\newblock IEEE Transactions on Games \textbf{11}(3) (2019)  227--237

\bibitem{Dota}
Font, J.M., Mahlmann, T.:
\newblock Dota 2 bot competition.
\newblock IEEE Transactions on Games \textbf{11}(3) (2019)  285--289

\bibitem{ViZDoom}
Wydmuch, M., Kempka, M., Jaśkowski, W.:
\newblock Vizdoom competitions: Playing doom from pixels.
\newblock IEEE Transactions on Games \textbf{11}(3) (2019)  248--259

\bibitem{aiwc}
Hong, C., Jeong, I., Vecchietti, L.F., Har, D., Kim, J.H.:
\newblock Ai world cup: Robot-soccer-based competitions.
\newblock IEEE Transactions on Games \textbf{13}(4) (2021)  330--341

\bibitem{fira}
{FIRA Roboworld Cup Official Website}:
\newblock (2020) \url{https://firaworldcup.org/3}, Accessed: August 13, 2022.

\bibitem{robocup}
{RoboCup Federation Official Website}:
\newblock (2020) \url{https://www.robocup.org/}, Accessed: August 13, 2022.

\bibitem{inbook}
Akiyama, H., Nakashima, T., Fukushima, T., Zhong, J., Suzuki, Y., Ohori, A.
\newblock In: HELIOS2018: RoboCup 2018 Soccer Simulation 2D League Champion.
  (08 2019)  450--461

\bibitem{Fukushima}
Fukushima, T., Nakashima, T., Akiyama, H.:
\newblock Evaluation-function modeling with neural networks for robocup soccer.
\newblock Electronics and Communications in Japan \textbf{102} (12 2019)

\bibitem{git}
{AI World Cup Simulation Environment}:
\newblock (2019) \url{https://github.com/aiwc/test_world}, Accessed: August 13,
  2022.

\bibitem{aiworldcup}
{KAIST 2020 AI World Cup}:
\newblock (2020) \url{http://aiworldcup.org/ai_soccer}, Accessed: August 13,
  2022.

\bibitem{ppo}
Schulman, J., Wolski, F., Dhariwal, P., Radford, A., Klimov, O.:
\newblock Proximal policy optimization algorithms (2017)

\bibitem{sac}
Haarnoja, T., Zhou, A., Abbeel, P., Levine, S.:
\newblock Soft actor-critic: Off-policy maximum entropy deep reinforcement
  learning with a stochastic actor (2018)

\bibitem{ddpg}
Lillicrap, T.P., Hunt, J.J., Pritzel, A., Heess, N., Erez, T., Tassa, Y.,
  Silver, D., Wierstra, D.:
\newblock Continuous control with deep reinforcement learning (2015)

\bibitem{wcg}
{World Cyber Games}:
\newblock (2019) \url{https://www.wcg.com/}, Accessed: May 15, 2020.

\bibitem{Youtube}
{WCG 2019 New Horizons | AI Masters Group Stage | amista vs CY95}:
\newblock (2019) \url{https://www.youtube.com/watch?v=nUqM2jiPYYE}, Accessed:
  August 13, 2022.

\end{thebibliography}

\end{document}